\documentclass[10pt,twocolumn,letterpaper]{article}

\usepackage{authblk}

\makeatletter
\renewcommand\AB@affilsepx{~ ~~\protect\Affilfont}
\makeatother

\usepackage[pagenumbers]{cvpr} 

\usepackage{graphicx}
\usepackage{amsmath}
\usepackage{amssymb}
\usepackage{booktabs}
\usepackage{xcolor} 

\usepackage{multirow}

\usepackage[pagebackref,breaklinks,colorlinks]{hyperref}

\usepackage[capitalize]{cleveref}
\crefname{section}{Sec.}{Secs.}
\Crefname{section}{Section}{Sections}
\Crefname{table}{Table}{Tables}
\crefname{table}{Tab.}{Tabs.}

\definecolor{Brown}{rgb}{0.55,0.27,0.1}

\newcommand{\y}{\mathcal{Y}}
\newcommand{\s}{\mathrm{s}}
\newcommand{\tar}{\mathrm{t}}
\newcommand{\p}{\mathcal{P}}

\def\cvprPaperID{10074} 
\def\confName{CVPR}
\def\confYear{2022}

\begin{document}

\title{Domain Adaptation on Point Clouds via Geometry-Aware Implicits}


\author[1]{
	Yuefan Shen
}
\author[2]{
	Yanchao Yang
}
\author[3]{
	Mi Yan
}
\author[3]{
	He Wang
}
\author[1]{
	Youyi Zheng
}
\author[2]{
	Leonidas Guibas
}
\affil[1]{
	Zhejiang University
	\thanks{The authors from Zhejiang University are affiliated with the State Key Lab of CAD\&CG.}
}\affil[2]{
	Stanford University
}\affil[3]{
	Peking University
}

\maketitle

\begin{abstract}
As a popular geometric representation, point clouds have attracted much attention in 3D vision, leading to many applications in autonomous driving and robotics. One important yet unsolved issue for learning on point cloud is that point clouds of the same object can have significant geometric variations if generated using different procedures or captured using different sensors. These inconsistencies induce domain gaps such that neural networks trained on one domain may fail to generalize on others. A typical technique to reduce the domain gap is to perform adversarial training so that point clouds in the feature space can align. However, adversarial training is easy to fall into degenerated local minima, resulting in negative adaptation gains. Here we propose a simple yet effective method for unsupervised domain adaptation on point clouds by employing a self-supervised task of learning geometry-aware implicits, which plays two critical roles in one shot. First, the geometric information in the point clouds is preserved through the implicit representations for downstream tasks. More importantly, the domain-specific variations can be effectively learned away in the implicit space. We also propose an adaptive strategy to compute unsigned distance fields for arbitrary point clouds due to the lack of shape models in practice. When combined with a task loss, the proposed outperforms state-of-the-art unsupervised domain adaptation methods that rely on adversarial domain alignment and more complicated self-supervised tasks. Our method is evaluated on both PointDA-10 and GraspNet datasets. The code and trained models will be publicly available.
\end{abstract}

\section{Introduction}
\label{sec:intro}

\begin{figure}[!t]
	\vspace{-0.2cm}
	\centering
	\includegraphics[width=1.0\linewidth]{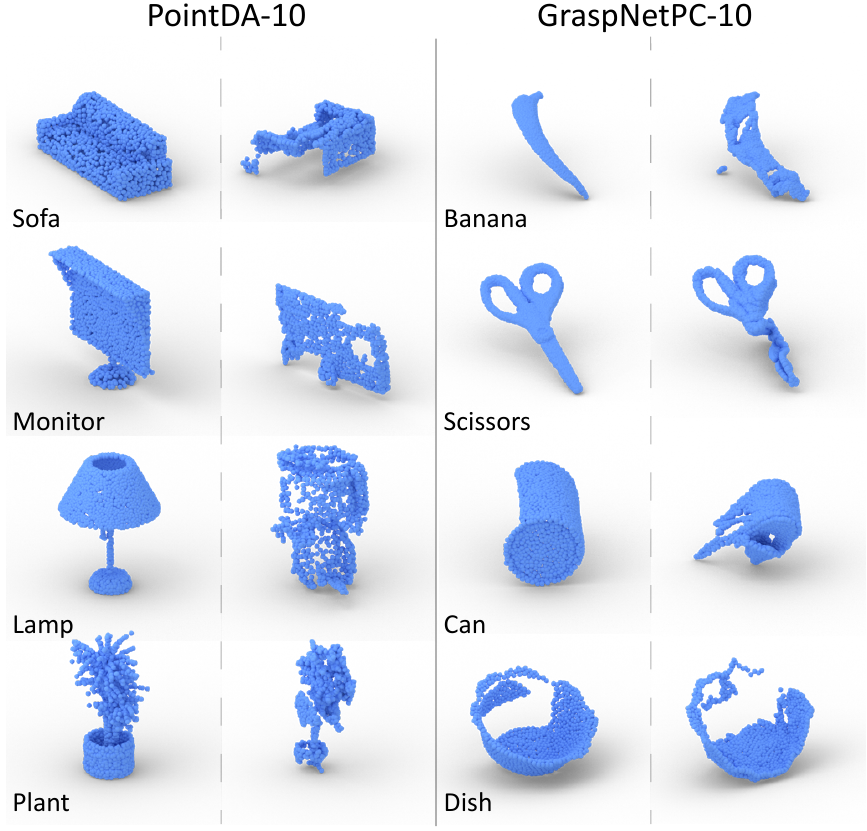}
	\caption{Point clouds in the real world exhibit diverse geometric variations caused by differences in the data capture pipeline. Given these variations, networks trained on one collection of point clouds may incur a performance drop when tested on different ones. Thus adaptation is needed to alleviate generalization issues, especially for domains where the annotation is scarce.}
	\vspace{-0.6cm}
	\label{fig:teaser}
\end{figure}

Point clouds captured under different settings can exhibit prominent variations that cause performance drop when neural networks are tested on a domain that is different from the training ones.
This can be troublesome if the network can {\it not} be fine-tuned due to time constraints or limited computational budget. 
More often, labels needed for fine-tuning on the test domain are simply unavailable due to high annotation cost, which is the situation we are interested in and is always formulated as unsupervised domain adaptation (UDA) problems.
In UDA, the source domain comes with rich annotations, while the target domain has no annotation at all.
The {\it key} to a successful domain adaptation lies in two folds. First, the two domains have to be (statistically) aligned, either in the point cloud space or in a feature space, so that the shared mapping to the output space can now operate on the same ground across domains.
Moreover, the alignment between domains has to be semantically meaningful, e.g., chairs in the source should be aligned with chairs in the target.
Otherwise, the shared mapping can still fail in predicting the labels even if the two domains are aligned.

Existing UDA methods on point clouds mainly rely on two mechanisms to align the domains.
One is to perform domain adversarial training and enforce the features of point clouds from both domains to be indistinguishable by domain discriminators.
Since adversarial training is unstable and easy to get stuck at degenerated local minimas, 
there is little guarantee that the alignment would be semantically meaningful.
For example, adversarial training could distort the geometric information in the point clouds by eliminating too many variations while aligning the domains.
In this case, the alignment can result in negative adaptation gains.
An extra layer of difficulty is that the alignment process could be highly sensitive to the architecture of the discriminators for point clouds as shown in~\cite{wang2020rethinking}, thus making the alignment more uncontrollable.

The other mechanism is to perform domain alignment through learning self-supervised tasks.
The underlying motivation is that a well-designed self-supervised task can facilitate learning domain invariant features since the task itself is shared across domains.
A diverse set of carefully designed self-supervised tasks are proposed, which focus on predictive tasks where the self-supervised labels are generated by
augmenting or modifying the original point clouds.
For instance, rotation angle classification~\cite{zou2021geometry} and deformation regression~\cite{achituve2021self}. 
Compared to domain adversarial training,
self-supervised learning enables explicit control over the invariants been learned by adjusting the self-supervised tasks.
Consequently, one can also regularize the alignment process through this knob.

We take the latter approach, 
but we resort to a self-supervised task 
where the supervision comes from the point clouds themselves, instead of manually designed classification labels.
Specifically, we ask for a latent space that encodes the underlying geometry of the point clouds through implicit functions.
As the geometry is explicitly modeled and preserved,
these latents or implicits should maintain sufficient information for the main task and help prevent mismatch in semantics caused by distortions during the alignment.
Due to the lack of shape models,
we propose an adaptive unsigned distance field that enables training the implicits for arbitrary point clouds, especially for the ones that are sparse and irregularly sampled.
After the initial round of adaptation, we follow the literature and apply self-training with pseudo labels in the target domain to further close the gap. 
We experiment on two major point cloud datasets, PointDA-10~{qin2019pointdan} and GraspNet~{fang2020graspnet}, to report the performance of the proposed method and evaluate the effectiveness of each component. Our contributions are:
\begin{itemize}
    \item The first method leverages implicit function learning as a self-supervised task for unsupervised domain adaptation on point clouds.
    \vspace{-0.1cm}
    \item Effective training strategies to make our method robust to diverse artifacts exhibited in the point clouds.
    \vspace{-0.1cm}
    \item State-of-the-art performance on two major datasets, PointDA-10~\cite{qin2019pointdan} and GraspNet~\cite{fang2020graspnet}. Moreover, we are the first to report results on GraspNet.
\end{itemize}


\begin{figure*}[!t]
	\centering
	\includegraphics[width=1.0\linewidth]{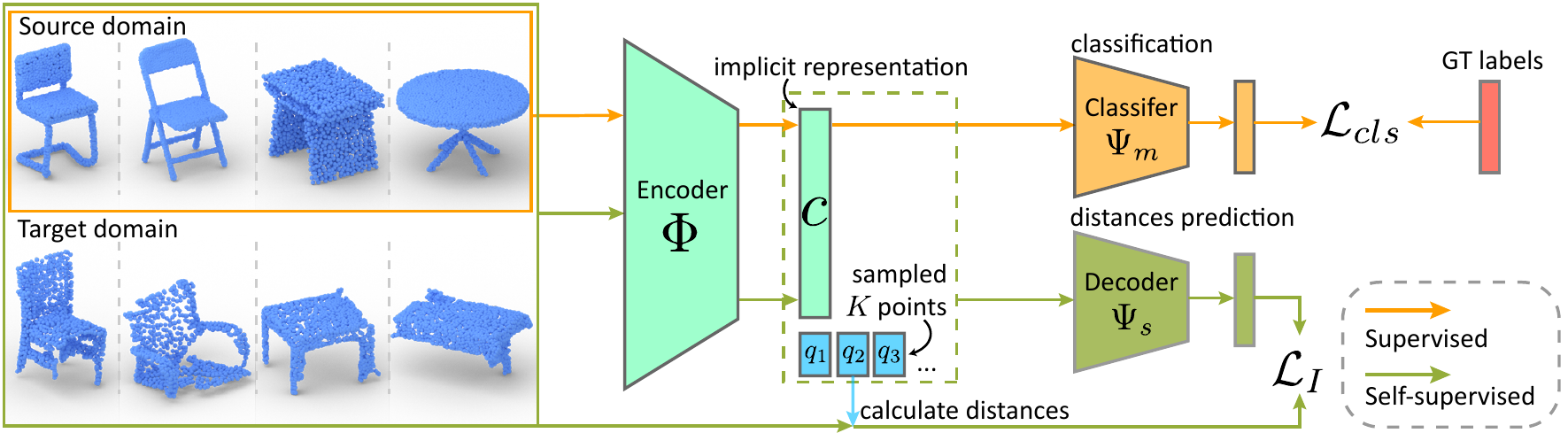}
	\vspace{-0.5cm}
	\caption{Overview of the proposed framework for unsupervised domain adaptation on point clouds. The two pathways (supervised and self-supervised) in our framework are marked with different colors. The supervised pathway takes as input the point clouds from the source domain and calculates the cross-entropy loss with ground-truth labels. The self-supervised pathway takes point clouds from the source and target domains and calculates the self-supervised loss with the proposed adaptive unsigned distances between sampled points and the input point clouds. Note, in the self-paced self-training stage, the classifier is also trained with pseudo labels.
	}
	\vspace{-0.6cm}
	\label{fig:framework}
\end{figure*}

\section{Related Work}

\subsection{Deep Learning on Point Clouds}

To handle the irregularity and permutation-invariance of point clouds, various methods have been proposed. PointNet~\cite{qi2017pointnet} and PointNet++\cite{qi2017pointnet++} use max-pooling as a permutation-invariant local feature extractor and the latter gathers local features in a hierarchical way. DGCNN~\cite{wang2019dynamic} considers a point cloud as a graph and dynamically updates the graph to aggregate features. Recently, Point Transformer~\cite{zhao2021point} adopts transformer for point cloud processing which achieves state-of-the-art performance in several benchmarks.

\subsection{2D and 3D Unsupervised Domain Adaptation}


Extensive works have been proposed to perform UDA on 2D images, which can be classified into two categories, i.e., the methods based on domain-invariant feature learning and methods for learning domain mapping. The former ones~\cite{kang2019contrastive,long2018transferable,rozantsev2018beyond,ganin2016domain,tzeng2017adversarial,saito2018maximum} minimize the discrepancy between two distributions in the feature space, while the latter ones~\cite{shrivastava2017learning,bousmalis2017unsupervised,hoffman2018cycada} directly learn the translation from the source domain to the target domain using neural networks, e.g., CycleGAN ~\cite{zhu2017unpaired}. Despite their differences, domain adversarial training is widely exploited in these methods. Several useful techniques are also proposed, for example, pseudo-labeling~\cite{saito2017asymmetric}, and batch normalization tailored for domain adaptation~\cite{maria2017autodial}.



Though lots of efforts have been made on 2D images, UDA on 3d point cloud is still in its early stage. As discussed in the introduction, UDA on point clouds can be roughly divided into two categories. The first category~\cite{qin2019pointdan} directly extends domain adversarial training used in 2D images to 3D point clouds to align features on both local and global levels. However, unlike previous works on the 2D domain, adversarial methods on 3D point clouds can not balance well between local geometry alignment and global semantic alignment. Most recent works in UDA on point clouds fall in the second category, i.e., focusing on designing suitable self-supervised tasks on point clouds to facilitate learning domain invariant features, which we discuss in detail in the following subsection.

Apart from UDA on object point clouds, several methods are proposed to address specific domain gaps on LiDAR point clouds, where the common factors are depth missing and sampling difference between sensors. Both~\cite{zhao2020epointda} and~\cite{saleh2019domain} use CycleGAN~\cite{zhu2017unpaired} to generate more realistic LiDAR point clouds from synthetic data, i.e., sim2real. Complete \& Label~\cite{yi2021complete} leverages segmentation on completed surface reconstructed from sparse point cloud for better adaptation.

\subsection{Self-Supervised Learning on Point Clouds}
\label{subsec:SSL}

Previous works design various kinds of self-training tasks to align the two domains. GAST~\cite{zou2021geometry} proposes rotation classification and distortion localization as a self-supervised task to align features at both local and global levels. DefRec~\cite{achituve2021self} proposes deformation-reconstruction and~\cite{luo2021learnable} extends it into a learnable deformation task to further improve the performance. RS~\cite{sauder2019self} shuffles and restores the input point cloud to improve discrimination.

However, there are two main issues with these methods. Some of them can not be applied to more challenging datasets where object point clouds are not aligned and are heavily occluded, resulting in ambiguity in the rotation prediction~\cite{zou2021geometry,poursaeed2020self} and restoring~\cite{sauder2019self} tasks. Besides, by aligning high-level features~\cite{zou2021geometry,achituve2021self,luo2021learnable,sauder2019self}, i.e., in semantic space, they could lose valuable information of the underlying geometry, which limits their applicability to more general geometric processing tasks. Motivated by these two observations, we design a task where the point cloud itself generates the self-supervision on the two domains and features are aligned to preserve geometric primitives. The aligned features can further be used for high-level semantic extraction, making our method more general for various main tasks.


\section{Method}\label{sec:method}

We tackle unsupervised domain adaptation (UDA) on point clouds for classification.
Let $\mathcal{P} \in \mathbb{R}^{N \times 3}$ be a point cloud consisting of the spatial coordinates of $N$ points in the 3D space.
Accordingly,
let $\mathcal{D}^{\s}=\{\mathcal{P}_i^{\s}, \y_i^\s\}$ be the point clouds and their ground-truth labels from the source domain.
Similarly,
$\mathcal{D}^{\tar}=\{\mathcal{P}_i^{\tar}\}$ is the collection of target domain point clouds whose labels are missing.
Our goal is to train a network $\Theta$, i.e., $\y=\Theta(\p)$ using the labeled point clouds from the source domain so that it can work well on the target point clouds without further labeling.

The {\it key} is to {\it align} the point clouds from both domains, and at the same time, ensure that the correspondence is {\it semantically meaningful}, i.e., the point clouds of the same category are expected to be aligned after the adaptation. 
One can apply domain adversaries for aligning domains, however, the alignment is hard to control and may result in negative adaptation gains due to difficulties in adversarial training.
We resort to the strategy of utilizing self-supervised tasks
that are shared across domains for alignment in a multi-task fashion.
This enables an explicit control of the meaningfulness of the alignment by selecting an appropriate self-supervised task.
There are two pathways in our framework, as shown in Fig.~\ref{fig:framework}.
The {\it main task} is performed by $\Phi$ and $\Psi_m$, i.e., $\Theta=\Psi_m\circ\Phi$, with $\Phi$ an encoder that extracts features from the point clouds and $\Psi_m$ the main task head (classifier).
Likewise, the {\it self-supervised task} is performed by $\Phi$ (shared with the main task pathway) and $\Psi_s$, which can be trained on both domains.
Next, we detail each of the proposed components and their training.

\subsection{Self-Supervised Geometry-Aware Implicit} \label{subsec:imp}

Implicit representations are capable of preserving complex details for given shapes~\cite{park2019deepsdf,chen2019learning}.
Instead of high-quality shape reconstruction, 
we leverage the implicit representation space for aligning point clouds from different domains by performing the following self-supervised task.

Given a point cloud $\mathcal{P}$, either complete or partial, 
the shared encoder $\Phi$ first maps it to a feature vector
$c = \Phi(\mathcal{P})$ as the implicit representation of the unknown underlying shape from where $\p$ is observed. 
Suppose $\mathcal{Q} \in \mathbb{R}^{K \times 3}$ are $K$ randomly sampled points in the unit cube.
By definition,
the implicit value (e.g., distance to the surface)
for each point $q \in \mathcal{Q}$ can be decoded as:
\begin{equation}
	f_{\p}(q) = \Psi_s(q, c)
	\label{eq:imp_decoder}
\end{equation}
where $f_{\p}$ is the implicit function of the underlying geometry conditioned on the input point cloud $\p$. 
Following the literature~\cite{chen2019learning,mescheder2019occupancy},
the decoder $\Psi_s$ takes as input the concatenation of the query point and the encoded implicit representation. 
Since the point clouds can be partial, 
we set the implicit values as unsigned distances to the underlying surface.
The computation of these values is described in the following.

\subsubsection{Adaptive Unsigned Distance of Point Cloud} \label{subsec:udf}

\begin{figure}[!t]
	\centering
	\includegraphics[width=1.0\linewidth]{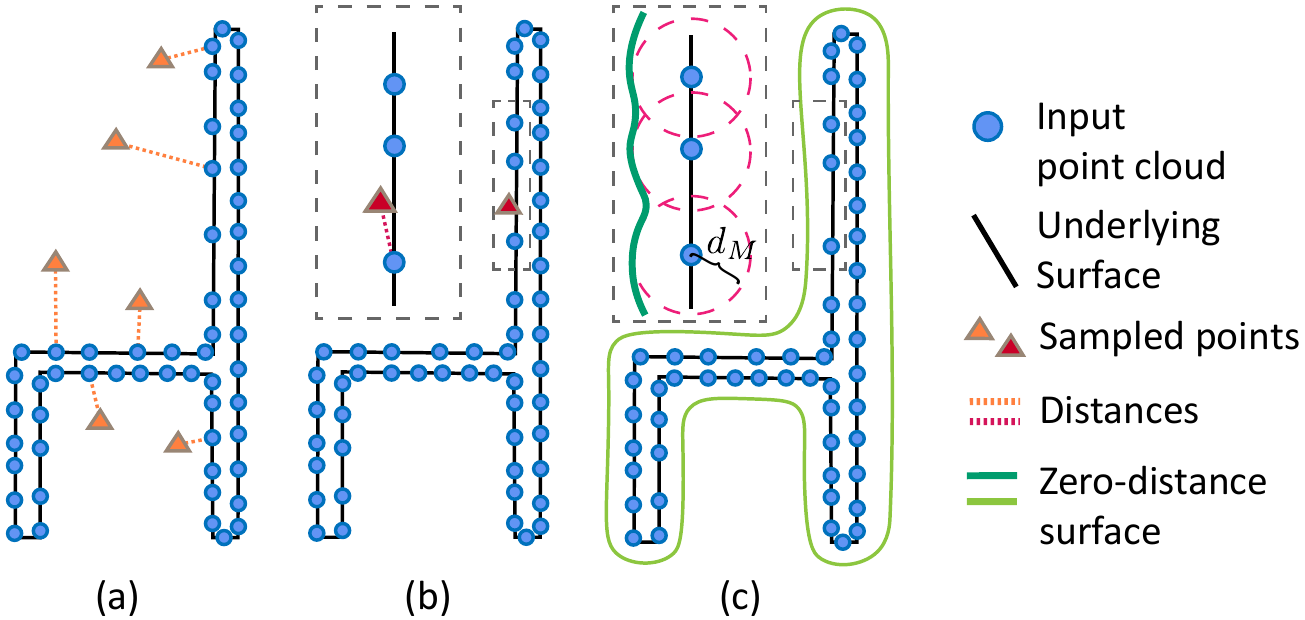}
	\caption{Adaptive unsigned distance field. (a): examples of calculating distances from sampled points (triangle) to their nearest points in the input point cloud. (b): when a sampled point is close to the surface, its nearest neighbor distance is still large due to sparsity. (c): the adaptive unsigned distance field and the zero-surface, with $d_M$ the adaptive clamping value.}
	\label{fig:adaptive-distance}
\end{figure}

Different from reconstruction where the known meshes can be used to compute the ground-truth for the distance values, we only have access to the point clouds.
However, as our goal is to leverage the implicit representation to align domains and reduce performance drop,
we do not need the implicits to perfectly represent the underlying geometry and reconstruct the point cloud.
To this end, we can compute approximates of the unsigned distance fields to supervise the training of the implicit space.

An intuitive method is to approximate the unsigned distance from a query point to the underlying surface by the distance between the same query point and its nearest neighbor from the point cloud (Fig.~\ref{fig:adaptive-distance} (a)).
This could work if the point clouds are densely and uniformly sampled.
Nevertheless, in practice, point clouds are usually sparse and irregularly sampled due to sensor noise and complex geometry in the scene.
These peculiarities can cause problems for the nearest neighbor approximations. 
For example, 
when the query point is very close to the underlying surface, the distance could still be large, as shown in Fig.~\ref{fig:adaptive-distance} (b). 
Thus the learned implicit space may not faithfully represent the geometry of the point clouds and can induce performance drop across domains.

To prevent unexpected distortions of the geometry in the approximation,
we propose an adaptive clamping technique based on a global average over statistics of the local geometry. For a point $p_j$ in the input point cloud $\mathcal{P}$, 
we first calculate the mean of the distances between $p_j$ and its $M$ nearest neighbors within $\p$:
\begin{equation}\label{eq:self_dis}
	d_j = \frac{1}{M}\sum_m\|p_m - p_j\|
\end{equation}
where $p_m$ is from the $M$ nearest neighbours and we name $d_j$ the {\it local affinity} of point $p_j$.
Then, we compute the average of the local affinity of all the points in the point cloud, i.e.,
$d_M = \frac{1}{N}\sum_{j=0}^{N-1}{d_j}$, which is the {\it adaptive clamping threshold} 
and is used in the following to compute the adaptive approximate of the unsigned distance field from the point cloud:
\begin{equation}\label{eq:soft_df}
	d_\p(q) =
	\begin{cases}
		\|q - p^*(q)\|   & \quad \text{if } \|q - p^*(q)\| > d_M\\
		0  			& \quad \text{otherwise}
	\end{cases}
\end{equation}
where $p^*(q)$ is the nearest neighbor of the query $q$ in the point cloud $\p$.
Also, note that $d_M$ depends on $\p$ and can vary between point clouds to accommodate different sparsity levels.
An example of the adaptive unsigned distance field can be found in Fig.~\ref{fig:adaptive-distance} (c).
As observed, the unsigned distance field approximated via Eq.~\eqref{eq:soft_df} captures the underlying geometry of the point cloud and is more robust to sampling issues.
With the adaptive unsigned distance field $d_\p$,
the self-supervised loss for learning the implicit space is:
\begin{equation}\label{eq:sl_loss}
	\mathcal{L}_I = \frac{1}{|\mathcal{Q}|}\sum_{q\in\mathcal{Q}}|f_\mathcal{P}(q) - d_\p(q)|
\end{equation}
here, $|\mathcal{Q}|$ is the cardinality of the sampled query points.
Next, we discuss a few issues encountered during the articulation of the whole pipeline and our solutions.

\subsubsection{Point Cloud Augmentation} \label{subsec:datapre}

\begin{figure}[!t]
	\centering
	\includegraphics[width=1.0\linewidth]{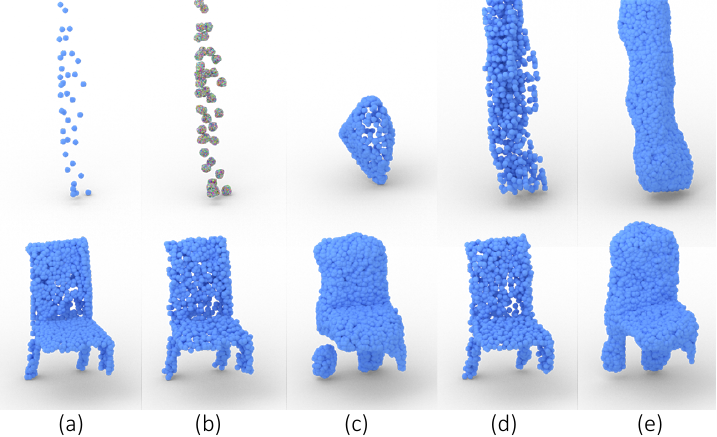}
	\caption{(a): Input point clouds with duplication. The point cloud in the top row has only 38 unique points, 
	but it contains 1024 points in total with duplication to account for the fixed input size of the backbone. 
	(b): Point clouds after random jittering in the range $[0.03, 0.06]$ (instead of duplicating points), which simply dilates each point. The points in the top row are colored randomly for better discrimination.
	(c): Point clouds sampled from the implicits learned with the random jittering scheme in (b). We can observe much heavier geometric distortion for sparser point clouds (top vs. bottom).
	(d): Point clouds jittered using the proposed scheme based on the spatially varying local affinity measure $d_j$ in Eq.~\eqref{eq:self_dis}. 
	(e) Sampled point clouds from implicits learned with the jittering scheme in (d), which preserves the geometry for both sparse and dense point clouds.}
	\label{fig:jitter}
\end{figure}

{\bf Jittering.} The point cloud backbone usually assumes a fixed number of points during training, for example, 1024 points for a single point cloud.
However, in practice, the number of points in a single point cloud may not be the same due to irregular sampling or different shape sizes.
For example, in the unsupervised domain adaptation benchmark PointDA-10~\cite{qin2019pointdan}, point clouds from ModelNet and ScanNet can have very different numbers of points.
A commonly used technique is to pad the point clouds to the same number of points through jittering, which may also improve the training if the jittering is properly designed.
The simplest jittering method is to add duplicate points to the original point cloud (Fig.~\ref{fig:jitter} (a)).
Another method is to add uniform random perturbations (Fig.~\ref{fig:jitter} (b)).
However, both methods will generate points that make the local affinity measure uninformative, so that the proposed adaptive unsigned distance field may not be effective in preventing geometric distortions for sparse and irregular point clouds.
As shown in Fig.~\ref{fig:jitter} (c), the resampled point clouds from the learned implicits with random perturbations exhibit significant geometric distortions for sparse point clouds.

In order to avoid such degenerated cases in the padding procedure,  we propose to perform the point jittering in an affinity-aware manner.
Similar to calculating the adaptive unsigned distance field,
for each point $p_j$ in the raw point cloud, 
we first obtain its local affinity $d_j$ using Eq.~\eqref{eq:self_dis}.
A random offset in the range of $[-\frac{d_j}{2}, \frac{d_j}{2}]$ is then added to $p_j$ to generated jittered points. 
The point clouds generated with this jittering scheme have little deviation from the raw point clouds, as shown in Fig.~\ref{fig:jitter} (d).
Moreover, the resampled point clouds from the implicits learned with the affinity-aware jittering maintain the underlying geometry for both sparse and dense point clouds as observed in Fig.~\ref{fig:jitter} (e).

{\bf Random masking.}
Point clouds may come in a partial form due to self-occlusions.
To improve the robustness of the implicits concerning the partiality and further reduce the domain variations,
we choose to mask out a local neighborhood of a randomly selected point as an additional data augmentation.

Let $\mathcal{P}$ be a point cloud before random masking, and $\hat{\mathcal{P}}$ be the point cloud obtained by dropping out a neighborhood of radius of $r_m$. 
We ask the implicits of both point clouds to be similar as they are sampled from the same geometry.
During training, we add a loss term between the implicit representations of the input point cloud and its masked version:
\begin{equation}\label{eq:mask_loss}
	\mathcal{L}_M = \|\Phi(\mathcal{P}) - \Phi(\hat{\mathcal{P}})\|
\end{equation}
with $\|\cdot\|$ the L-2 distance.

\subsection{Self-Paced Self-Training} \label{subsec:spsl}

The main task we tackle here is point cloud classification.
Before adaptation, we only have labeled data in the source domain, i.e., $\{\mathcal{P}_i^{\s}, \y_i^\s\}$, which allows us to train the main task branch with a cross-entropy loss:
\begin{equation}
	\mathcal{L}_{cls}^\s = -\frac{1}{N_{\s}}\sum_{i=1}^{N_{\s}}\sum_{j=1}^{J}\y_{i, j}^\s\\log(\Psi_m(\Phi(\mathcal{P}_i^{\s}))_j)
\end{equation}
where $\y_{i, j}^\s$ represents the ground-truth one-hot labels and $\Psi_m(\Phi(\mathcal{P}_i^{\s}))_j$ is the predicted probability for the $j^{th}$ class.

When the initial adaptation is made,
point clouds from the source and target domains should be aligned to some extent.
In this case, techniques used in semi-supervised learning are now in their functioning state.
For example, GAST~\cite{zou2021geometry} employs the strategy of self-paced self-training (SPST)~\cite{lee2013pseudo,zou2018unsupervised} to further align the two domains by generating pseudo labels in the target domain using highly confident predictions. 
We follow this strategy to squeeze more juice out of the source labels. 
Let $\hat{\y}_i^\tar$ be the predicted pseudo labels, the loss function to perform the self-training is:
\begin{equation}\label{eq:sp_loss}
	\mathcal{L}_{cls}^\tar = -\frac{1}{N_{\tar}}\sum_{i=1}^{N_{\tar}}(\sum_{j=1}^{J}\hat{\y}_{i, j}^\tar\\log(\Psi_m(\Phi(\mathcal{P}_i^{\tar}))_j) + \gamma|\hat{\y}_i^\tar|_1)
\end{equation}
Similarly, the first term in Eq.~\eqref{eq:sp_loss} is a cross-entropy loss between the target pseudo labels and the predictions, and the second term is used to avoid degenerate solutions that assign all $\hat{\y}^\tar$ as $0$. 
We follow~\cite{zou2021geometry,zou2018unsupervised} to apply a two-stage optimization using Eq.~\eqref{eq:sp_loss}, where the pseudo labels are first computed using nonlinear integer programming. Then the branch $\Psi_m\circ\Phi$ is updated using the pseudo labels.
These two steps are performed iteratively to adapt between the source and target domains progressively.
The hyper-parameter $\gamma$ controls the number of selected target samples.


\subsection{Overall Loss} \label{subsec:all}
The overall training loss of our method is:
\begin{equation}\label{eq:loss}
	\mathcal{L} = \mathcal{L}_I + \alpha\mathcal{L}_M + \beta\mathcal{L}_{cls}^\s + \mu\mathcal{L}_{cls}^\tar
\end{equation}
Note, the self-supervised implicit representation learning can be pre-trained on point clouds to encourage faster convergence during adaptation, i.e., set $\beta, \mu=0$. 
After pre-training the networks $\Phi\circ\Psi_s$ for learning geometry-aware implicits, together with loss terms of the classification task $\mathcal{L}_{cls}^\s$ and $\mathcal{L}_{cls}^\tar$ can be added back to perform the joint domain adaptation.

\section{Experiments}

\begin{table*}[!t]
	\centering
	\scalebox{0.88}{
		\begin{tabular}{lccc||ccccccc}
			\toprule
			\bf{Methods} & Adv. & SLT & SPST & M$\rightarrow$S & M$\rightarrow$S* & S$\rightarrow$M & S$\rightarrow$S* & S*$\rightarrow$M & S*$\rightarrow$S & Avg.\\
			\midrule
			Supervised                                  &            &            &            & 93.9 $\pm$ 0.2 & 78.4 $\pm$ 0.6 & 96.2 $\pm$ 0.1 & 78.4 $\pm$ 0.6 & 96.2 $\pm$ 0.1 & 93.9 $\pm$ 0.2 & 89.5 \\
			Baseline (w/o adap.)                                    &            &            &            & 83.3 $\pm$ 0.7 & 43.8 $\pm$ 2.3 & 75.5 $\pm$ 1.8 & 42.5 $\pm$ 1.4 & 63.8 $\pm$ 3.9 & 64.2 $\pm$ 0.8 & 62.2 \\
			\midrule
			DANN~\cite{ganin2016domain}                  & \checkmark &            &            & 74.8 $\pm$ 2.8 & 42.1 $\pm$ 0.6 & 57.5 $\pm$ 0.4 & 50.9 $\pm$ 1.0 & 43.7 $\pm$ 2.9 & 71.6 $\pm$ 1.0 & 56.8 \\
			PointDAN~\cite{qin2019pointdan}              & \checkmark &            &            & 83.9 $\pm$ 0.3 & 44.8 $\pm$ 1.4 & 63.3 $\pm$ 1.1 & 45.7 $\pm$ 0.7 & 43.6 $\pm$ 2.0 & 56.4 $\pm$ 1.5 & 56.3 \\
			RS~\cite{sauder2019self}                     &            & \checkmark &            & 79.9 $\pm$ 0.8 & 46.7 $\pm$ 4.8 & 75.2 $\pm$ 2.0 & 51.4 $\pm$ 3.9 & 71.8 $\pm$ 2.3 & 71.2 $\pm$ 2.8 & 66.0 \\
			DefRec+PCM~\cite{achituve2021self}         &            & \checkmark &            & 81.7 $\pm$ 0.6 & 51.8 $\pm$ 0.3 & 78.6 $\pm$ 0.7 & 54.5 $\pm$ 0.3 & 73.7 $\pm$ 1.6 & 71.1 $\pm$ 1.4 & 68.6 \\
			\multirow{2}{*}{GAST~\cite{zou2021geometry}} &            & \checkmark &            & 83.9 $\pm$ 0.2 & 56.7 $\pm$ 0.3 & 76.4 $\pm$ 0.2 & 55.0 $\pm$ 0.2 & 73.4 $\pm$ 0.3 & 72.2$\pm$ 0.2  & 69.5 \\
			                                            &            & \checkmark & \checkmark & 84.8 $\pm$ 0.1 & \textbf{59.8} $\pm$ 0.2 & 80.8 $\pm$ 0.6 & 56.7 $\pm$ 0.2 & 81.1 $\pm$ 0.8 & \textbf{74.9} $\pm$ 0.5 & 73.0 \\
			\midrule
			\multirow{2}{*}{Ours}                       &            & \checkmark &            & 85.8 $\pm$ 0.3 & 55.3 $\pm$ 0.3 & 77.2 $\pm$ 0.4 & 55.4 $\pm$ 0.5 & 73.8 $\pm$ 0.6 & 72.4 $\pm$ 1.0 & 70.0 \\
			                                            &            & \checkmark & \checkmark & \textbf{86.2} $\pm$ 0.2 & 58.6 $\pm$ 0.1 & \textbf{81.4} $\pm$ 0.4 & \textbf{56.9} $\pm$ 0.2 & \textbf{81.5} $\pm$ 0.5 & 74.4 $\pm$ 0.6 & \textbf{73.2} \\
			\bottomrule
		\end{tabular}
	}
	\vspace{-0.2cm}
	\caption{Classification accuracy (\%) averaged over 3 seeds ($\pm$ SEM) on the PointDA-10 dataset. 
	M: ModelNet, S: ShapNet, S*: ScanNet; $\rightarrow$ indicates the adaptation direction. 
	Adv.: adversarial domain alignment, SLT: self-learning tasks, and SPST: self-paced self-training.}
	\label{tab:pointda_res}
	\vspace{-0.2cm}
\end{table*}

To show that the implicits effectively encode geometries of point clouds and verify the importance of the proposed adaptive unsigned distance filed, we examine the implicit reconstructions in Sec.~\ref{subsec:implicit_res}.
To have a comprehensive understanding of both the effectiveness and limitations of the implicits learned from unconstrained point clouds for aligning the domains, we evaluate the whole pipeline for point cloud UDA on the classification task with two major datasets.
We report our results with and without self-paced self-training.

We compare to a list of recent state-of-the-art methods on unsupervised point cloud domain adaptation: DANN~\cite{ganin2016domain}, PointDAN~\cite{qin2019pointdan}, RS~\cite{sauder2019self}, DefRec+PCM~\cite{achituve2021self} and GAST~\cite{zou2021geometry}. 
In addition, we report the results obtained from the same network trained in a supervised manner on the target domain (``Supervised'', upper-bound). 
For reference, the network trained in the source domain but tested on the target domain without any adaptation is also included (``Baseline'', lower-bound).

\subsection{Datasets}

{\bf PointDA-10}~\cite{qin2019pointdan} consists of three widely-used datasets: ModelNet \cite{wu20153d}, ShapeNet \cite{chang2015shapenet} and ScanNet \cite{dai2017scannet}. 
All three datasets share the same ten categories (bed, table, sofa, chair, etc.). 
ModelNet contains 4183 training and 856 test samples, while ShapeNet contains 17378 training and 2492 test samples. ModelNet and ShapeNet are both sampled from 3D CAD models. 
Unlike these synthetic point cloud datasets, ScanNet consists of point clouds from scanned and reconstructed real-world scenes. 
There are 6110 training samples and 2048 test samples in ScanNet, and point clouds therein are usually incomplete because of occlusion by surrounding objects in the scene or self-occlusion in addition to realistic sensor noises. 
We follow the data preparation procedure used in \cite{qin2019pointdan,achituve2021self,zou2021geometry}. 
Specifically, all object point clouds in all datasets are aligned along the direction of gravity, while arbitrary rotations along the $z$ axis are allowed. 
Moreover, the input point cloud with batching is a list of 1024 points, which are sampled with duplicative padding from the original point clouds and are normalized to a unit scale.

\begin{figure}[!t]
	\centering
	\includegraphics[width=1.0\linewidth]{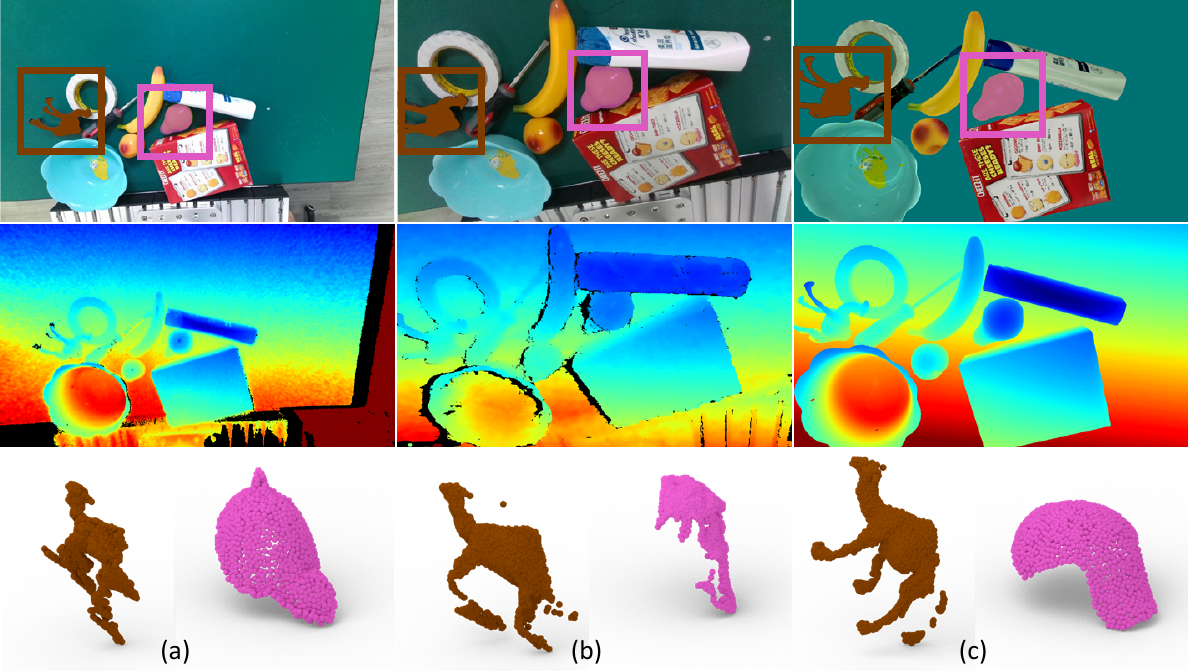}
	\caption{Point clouds from GraspNetPC-10 created with GraspNet~\cite{fang2020graspnet}. (a-b): RGBD and raw point cloud captured by Kinect and Realsense devices, respectively, and (c): Synthetic RGBD and point cloud. Segmentation masks are provided, as shown in the first row. The corresponding re-projected and cropped point clouds are visualized in the same color at the bottom.}
	\label{fig:graspnet}
\end{figure}

{\bf GraspNetPC-10} In order to test the domain adaptation on sim-to-real and real-to-real and check how the adaptation copes with different types of sensor noise,
we introduce GraspNetPC-10. 
It is created from GraspNet~\cite{fang2020graspnet} proposed for training robotic grasping on raw depth scans and corresponding reconstructed 3D CAD models of various objects. 
As shown in Fig.~\ref{fig:graspnet}, we create GraspNetPC-10 by re-projecting raw depth scans to 3D space and applying object segmentation masks to crop out the corresponding point clouds. Meanwhile, we synthesize similar senses with the same objects and render the synthetic depth scans to re-project synthetic point clouds. Different from point clouds in PointDA-10, point clouds in GraspNetPC-10 are {\it not aligned}.

Raw depth scans in GraspNet~\cite{fang2020graspnet} are captured by two different depth cameras, Kinect2 and Intel Realsense, so we have two domains of real-world point clouds. 
Following PointDA-10, we collect synthetic and real-world point clouds for ten object classes. 
In the synthetic domain, there are 12,000 training point clouds. 
In the Kinect domain, there are 10,973 training and 2,560 testing point clouds.
Similarly, in the  Realsense domain, there are 10,698 training and 2,560 testing point clouds. 
The real-world point clouds from the two devices are always corrupted by different noises, and there exist different levels of geometric distortions and missing parts, as observed in Fig.~\ref{fig:teaser} and Fig.~\ref{fig:graspnet}.

\begin{figure}[!t]
	\centering
	\includegraphics[width=1.0\linewidth]{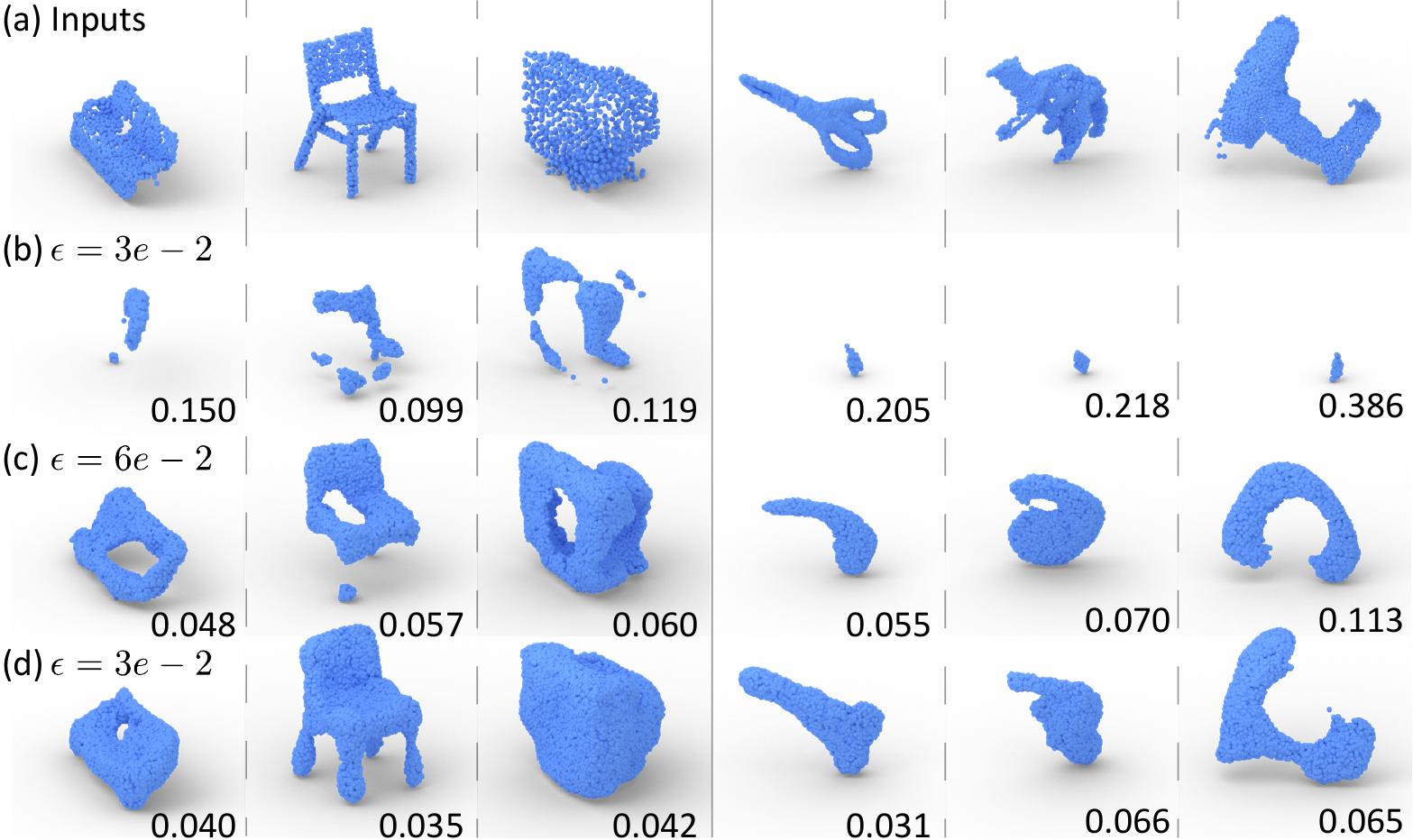}
	\caption{Visualization of resampled point clouds from the learned implicits. The left half shows test samples from the PointDA-10 dataset, and the right half shows test samples from GraspNetPC-10.
	(a): input point clouds, (b-c): resampled point clouds without adaptive unsigned distance field (AUD) at $\epsilon=$3e-2 and $\epsilon=$6e-2, (d): resampled point clouds with AUD at $\epsilon=$3e-2. The inserted numbers are the Chamfer distances between the resampled and the input point clouds (a).
	}
	\vspace{-0.3cm}
	\label{fig:recres}
\end{figure}

\subsection{Implementation Details}

Our experiments are conducted on servers with four GeForce RTX 3090 GPUs, and the networks are implemented within the PyTorch framework. For training, we use an Adam optimizer, with an initial learning rate $0.001$, weight decay $0.00005$, and an epoch-wise cosine annealing learning rate scheduler. We train all models for 200 epochs on PointDA-10 and 120 epochs on GraspNetPC-10 with a batch size of 32.

{\bf Network architecture} Following~\cite{zou2018unsupervised}, we choose the commonly used point cloud processing network DGCNN~\cite{wang2019dynamic} as the backbone for the encoder $\Phi$. 
The implicit decoder $\Psi_s$ and the category classifier $\Psi_m$ are multi-layer perceptrons (MLP) with fully connected layers. Decoder $\Psi_s$ is a four-layers MLP $\{512, 256, 128, 1\}$ followed by ReLU activation function (to make the output distance always positive) and classier $\Psi_m$ is a three-layers MLP $\{512, 256, 10\}$ in view of 10 semantic classes.

{\bf Hyper parameters} We set $M = 3$ for searching nearest neighbor points when calculating our adaptive unsigned distance field (AUD). The radius for the random masking $r_m$ is sampled from a uniform distribution in the range of [0.1, 0.3]. The weights of loss terms are set to $\alpha = 100, \beta=1.0, \theta=1.0$ and we adjust them slightly for better convergence on different datasets.

\begin{table*}[!t]
	\centering
	\scalebox{0.92}{
		\begin{tabular}{l@{\hspace{4.8ex}}c@{\hspace{4.8ex}}c@{\hspace{4.8ex}}c@{\hspace{4.8ex}}||c@{\hspace{4.8ex}}c@{\hspace{4.8ex}}c@{\hspace{4.8ex}}c@{\hspace{4.8ex}}c}
			\toprule
			\bf{Methods} & Adv. & SLT & SPST & Syn.$\rightarrow$Kin. & Syn.$\rightarrow$RS. & Kin.$\rightarrow$RS. & RS.$\rightarrow$Kin. & Avg.\\
			\midrule
			Supervised                                  &            &            &            & 97.2 $\pm$ 0.8 & 95.6 $\pm$ 0.4 & 95.6 $\pm$ 0.3 & 97.2 $\pm$ 0.4 & 96.4 \\
			Baseline (w/o adap.)                                    &            &            &            & 61.3 $\pm$ 1.0 & 54.4 $\pm$ 0.9 & 53.4 $\pm$ 1.3 & 68.5 $\pm$ 0.5 & 59.4 \\
			\midrule
			DANN~\cite{ganin2016domain}                  & \checkmark &            &            & 78.6 $\pm$ 0.3 & 70.3 $\pm$ 0.5 & 46.1 $\pm$ 2.2 & 67.9 $\pm$ 0.3 & 65.7 \\
			PointDAN~\cite{qin2019pointdan}              & \checkmark &            &            & 77.0 $\pm$ 0.2 & 72.5 $\pm$ 0.3 & 65.9 $\pm$ 1.2 & 82.3 $\pm$ 0.5 & 74.4 \\
			RS~\cite{sauder2019self}                     &            & \checkmark &            & 67.3 $\pm$ 0.4 & 58.6 $\pm$ 0.8 & 55.7 $\pm$ 1.5 & 69.6 $\pm$ 0.4 & 62.8 \\
			DefRec+PCM~\cite{achituve2021self}         &            & \checkmark &            & 80.7 $\pm$ 0.1 & 70.5 $\pm$ 0.4 & 65.1 $\pm$ 0.3 & 77.7 $\pm$ 1.2 & 73.5 \\
			\multirow{2}{*}{GAST~\cite{zou2021geometry}} &            & \checkmark &            & 69.8 $\pm$ 0.4 & 61.3 $\pm$ 0.3 & 58.7 $\pm$ 1.0 & 70.6 $\pm$ 0.3 & 65.1 \\
			&            & \checkmark & \checkmark & 81.3 $\pm$ 1.8 & 72.3 $\pm$ 0.8 & 61.3 $\pm$ 0.9 & 80.1 $\pm$ 0.5 & 73.8 \\
			\midrule
			\multirow{2}{*}{Ours}                       &            & \checkmark &            & 81.2 $\pm$ 0.3 & 73.1 $\pm$ 0.2 & 66.4 $\pm$ 0.5 & 82.6 $\pm$ 0.4 & 75.8 \\
			&            & \checkmark & \checkmark & \textbf{94.6} $\pm$ 0.4 & \textbf{80.5} $\pm$ 0.2 & \textbf{76.8} $\pm$ 0.4 & \textbf{85.9} $\pm$ 0.3 & \textbf{84.4} \\
			\bottomrule
		\end{tabular}
	}
	\vspace{-0.1cm}
	\caption{Classification accuracy (\%) averaged over 3 seeds ($\pm$ SEM) on GraspNetPC-10. Syn.: Synthetic domain, Kin.: Kinect domain, RS.: Realsense domain. Our models achieve the best performance over all settings.}
	\label{tab:graspnet_res}
\end{table*}

\subsection{Implicit Reconstruction} \label{subsec:implicit_res}

We show the resampled point clouds from learned implicit representations for analyzing the quality of the self-supervised geometry-aware implicits. Once the implicit encoder-decoder ($\Psi_s\circ\Phi$) is trained, given an input point cloud $\mathcal{P}$, we randomly sample $200,000$ points $q_s \in \mathbb{R}^{200000 \times 3}$ in the unit cube and calculate their unsigned distances conditioned on the original point cloud with our trained networks $d^{q_s} = \Psi_s(q_s, \Phi(\mathcal{P}))$. By setting a distance threshold $\epsilon$, we can choose the subset $\tilde{q_s} \subset q_s$ s.t. $d^{\tilde{q_s}} < \epsilon$ for visualization. If the distance field $f_{\mathcal{P}}$ is a good approximation of the underlying geometry, then $\tilde{q_s}$ will be similar to the input point clouds when $\epsilon$ varies.

In Fig.~\ref{fig:recres}, we compare our resampled point clouds with the input point clouds and the resampled point clouds from implicits learned without using our adaptive unsigned distance field, i.e., directly using the distances to the nearest neighbor but with a fixed clamping value. 
As observed, the resampled point clouds with AUD can preserve the underlying geometry well. However, using the same $\epsilon$, the resampled point clouds without using AUD (``w/o AUD'') are much inferior, meaning the learned implicits distort the geometry information. 
Moreover, we report the Chamfer distance between the resampled point clouds and the input.
Fig.~\ref{fig:recres} (c) shows the best resampled one for implicits learned without AUD (``w/o AUD'').
One can see that ``w/o AUD'' needs to apply a much larger $\epsilon$, i.e., two times larger than what is needed by ``AUD'', but the resampled point clouds are still severely deformed and exhibit lots of missing. 
These results demonstrate that our adaptive unsigned distance field is critical and effective in the proposed implicit representation alignment module.

\subsection{Unsupervised Domain Adaptation}

\begin{figure}[t]
	\centering
	\includegraphics[width=1.0\linewidth]{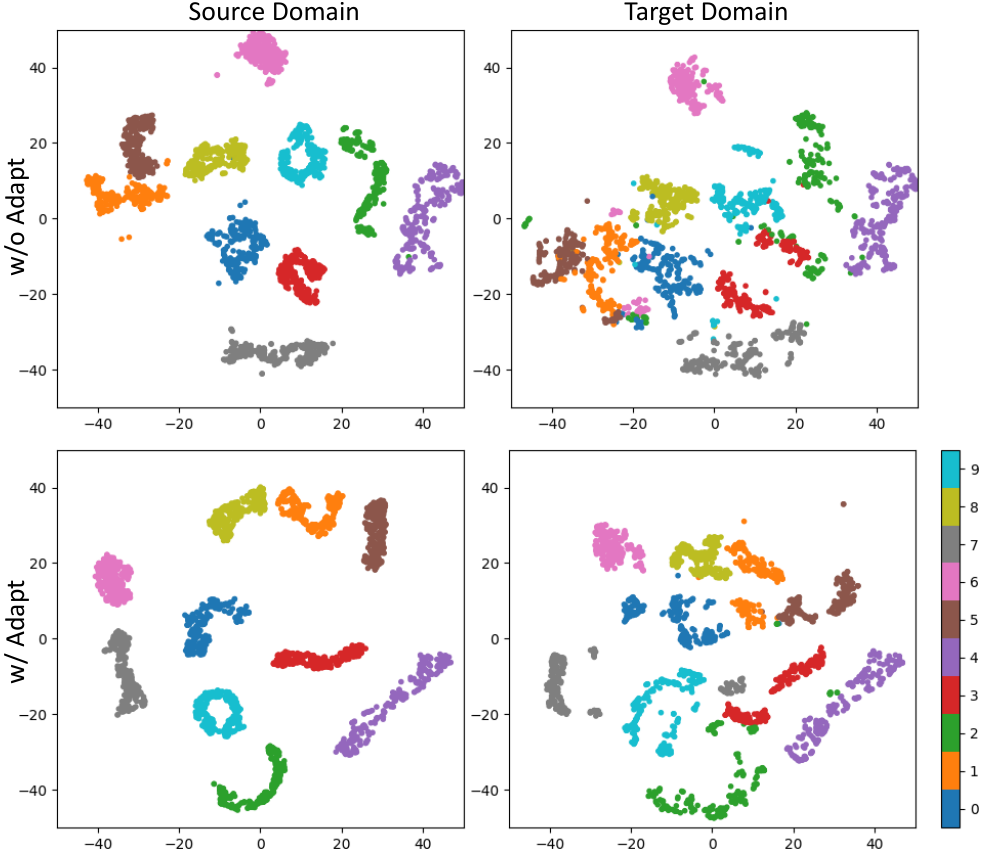}
	\caption{t-SNE~\cite{van2008visualizing} visualization of the output from our point cloud backbone on the Kinect domain (source) and the Realsense domain (target), which shows that the alignment through the implicits is effective, i.e., point cloud implicits from the target align better with the source ones after the adaptation (top vs. bottom). Different classes are displayed with different colors.}
	\label{fig:tsne}
\end{figure}

Table.~\ref{tab:pointda_res} and Table.~\ref{tab:graspnet_res} show the comparisons between our method and other state-of-the-art on PointDA-10 and GraspNetPC-10, respectively. 
For PointDA-10, we follow~\cite{zou2021geometry} and report performances on six different settings including ModelNet (M) $\leftrightarrow$ ShapeNet (S), M $\leftrightarrow$ ScanNet (S*) and S $\leftrightarrow$ S*. 
We find that methods utilizing self-learning tasks generally perform better than methods based on adversarial training, especially on ``synthetic to real'' settings. 
Compared to other self-leaning-based methods, our method (w/o SPST) excels on four settings and the average performance. After adding self-paced learning, our method competes with the most recent state-of-the-art method GAST~\cite{zou2021geometry} on PointDA-10. 
Compared to RS~\cite{sauder2019self} and DefRec+PCM~\cite{achituve2021self} that both use reconstruction-based self-learning tasks, our method again achieves better performance.

For GraspNetPC-10, our method outperforms the others with a significant margin before and after adding self-paced learning. One can observe a substantial decline of GAST~\cite{zou2021geometry} on GraspNetPC-10. It indicates that classification-based self-learning tasks may not be robust for different kinds of datasets. 
The local alignment method proposed in PointDAN~\cite{qin2019pointdan} now performs better than on the PointDA-10 dataset. DefRec+PCM~\cite{achituve2021self} ranks similarly. Our method achieves the highest score across all settings, whether with or without SPST. 
It is also worth noting that SPST is effective on all datasets, both GAST~\cite{zou2021geometry}, and our method improves with SPST. However, ``GAST+SPST'' is still worse than ours without SPST, which again shows the effectiveness of the proposed geometry-aware implicits for aligning domains with realistic sensor noise.

We also visualize the 1024-dimension latent codes in the implicit space using t-SNE~\cite{van2008visualizing}. As seen in Fig.~\ref{fig:tsne}, without domain adaptation, features of different classes in the target domain are mixed-up (e.g., class 1 and 5, class 2 and 3), and the overall distribution is different from that in the source domain. After adaptation, the distribution of the features in the target domain becomes similar to the source one and shows clear clusters. 

\section{Discussion}

It is challenging to align point clouds while maintaining a correct correspondence in terms of semantics without the target labels. 
However, we show that a simple alignment via the proposed implicit space training can be quite effective for the current unsupervised domain adaptation benchmarks on point clouds.
Our method achieves state-of-the-art performance on two benchmarks covering varying factors affecting the point cloud geometry within the data collection pipeline.
We hope our method can serve as a ground where low-level geometric distortions or variations are learned away so one can focus on high-level shape variations that are also generative factors for domain gaps.
This would require a carefully designed dataset with controllable disentangled elements of geometric variations and is out of the scope of our current work.

{\small
	\bibliographystyle{ieee_fullname}
	\bibliography{paper}
}

\end{document}